\documentclass[a4paper]{svproc}
\usepackage{url}

\usepackage[utf8]{inputenc}
\usepackage{graphics} 
\usepackage{epsfig} 
\usepackage{amsmath} 
\usepackage{amssymb}  
\usepackage{bm}
\usepackage{cite}
\usepackage[hyperfootnotes=false]{hyperref}
\usepackage{xcolor}
\usepackage{comment}
\allowdisplaybreaks
\usepackage{todonotes}
\usepackage{mathtools}
\usepackage{amsfonts}
\usepackage{float}
\usepackage{siunitx}
\usepackage{multirow}
\usepackage{algorithm} 
\usepackage{algpseudocode} 
\usepackage{graphicx}
\usepackage{subfig}
\usepackage[utf8]{inputenc}
\usepackage[font=footnotesize,labelfont=bf]{caption}
\usepackage{algorithm} 
\usepackage{algpseudocode} 
\usepackage{booktabs} 
\usepackage{tabularx}
    \newcolumntype{L}{>{\raggedright\arraybackslash}X}
\usepackage{lipsum}

\newcommand\blfootnote[1]{%
  \begingroup
  \renewcommand\thefootnote{}\footnote{#1}%
  \addtocounter{footnote}{-1}%
  \endgroup
}

\begin{document}
\mainmatter              
\title{Development of a semi-autonomous framework for NDT inspection with a tilting aerial platform}
\titlerunning{Tilting aerial manipulator}  
%
\author{Salvatore Marcellini \and Simone D'Angelo \and Alessandro De Crescenzo \and Michele Marolla \and Vincenzo Lippiello  \and Bruno Siciliano}
\authorrunning{Salvatore Marcellini et al.} 
%
\tocauthor{Salvatore Marcellini, Simone D'Angelo, Alessandro De Crescenzo, Michele Marolla, Vincenzo Lippiello}
%
\institute{{CREATE Consortium and PRISMA Lab, Department of Engineering and Information Technology, University of Naples Federico \uppercase\expandafter{\romannumeral2}, 80125, Naples, Italy}
\email{ \{salvatore.marcellini, simone.dangelo\}@unina.it}}

\maketitle              


\begin{abstract}
This letter investigates the problem of controlling an aerial manipulator, composed of an omnidirectional tilting drone equipped with a five-degrees-of-freedom robotic arm. The robot has to interact with the environment to inspect structures and perform non-destructive measurements. A parallel force-impedance control technique is developed to establish contact with the designed surface with a desired force profile. During the interaction, a pushing phase is required to create a vacuum between the surface and the echometer sensor mounted at the end-effector, to measure the thickness of the interaction surface. Repetitive measures are performed to show the repeatability of the algorithm.

\end{abstract}

\blfootnote{\small{This work has been supported by the AERO-TRAIN Project, European Union’s Horizon 2020 Research and Innovation Program under the Marie Skłodowska-Curie (Grant Agreement 953454), by the euROBIN project (Horizon 2020, Grant Agreement 101070596), and by AERIAL-CORE project (Horizon 2020, Grant Agreement No. 871479).}}
\vspace{-1.0cm}
\section{Motivation, Problem Statement, Related work}

Over the last decades, UAVs have been employed in different environments to help human operators or to autonomously accomplish tasks like surveillance~\cite{nmpc_surveillance}, industrial plants inspection~\cite{Cacace13, Aiello20, app11136220, Mao19} and similar. In recent years, thanks to the improvements in hardware devices and the solid aerial robotics community that provides new efficient control techniques, these robots have also been employed in tasks interacting with the environment. Among these are the non-destructive tests (NDT): this technique assesses the properties of materials, components, structures, or systems for inherent differences, welding flaws, or discontinuities without compromising the original part's integrity. Current NDT methods encompass a range of approaches in which some dedicated sensors need to remain in contact with the surface on which the measurement is to be made. 
In contrast, destructive testing alters or damages a component to such an extent that, even if it successfully passes the examination, it becomes unsuitable for further use.\\
Recently, several robotic systems devoted to executing NDT measurements have been developed; some solutions consider mobile platforms~\cite{SCHMIDT20131288} or internal climbing robots as in~\cite{7759332, Dertien11, 6907610} to navigate the interior part of structures such as tanks, pipes, and steam chests. Other climbing robots can adhere to the facility's external surface as in~\cite{IR, Ross03, Resino06}. 
\begin{figure}[t]
\centering
\subfloat[]{\includegraphics[height=0.25\textwidth]{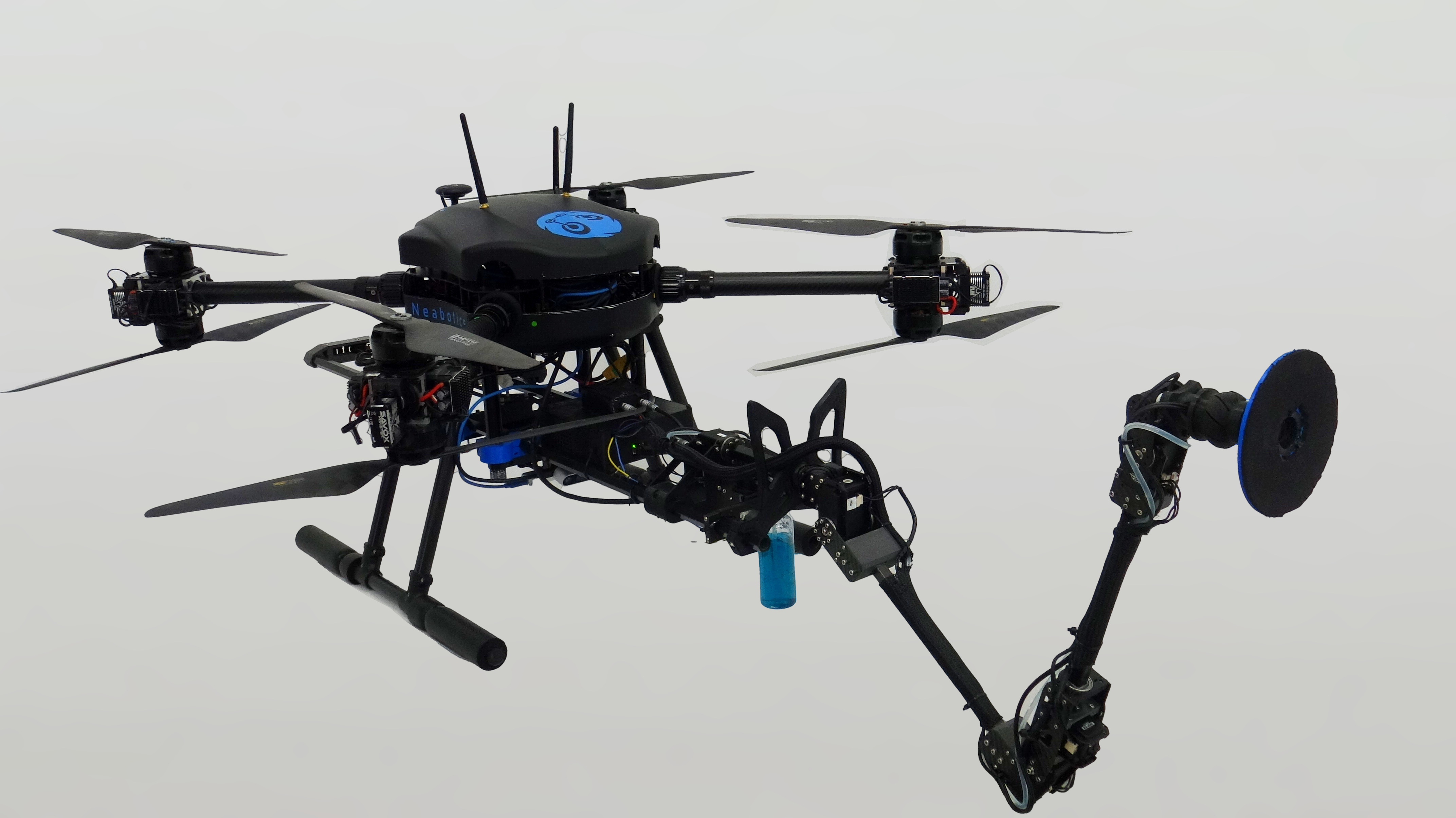}} \,
\subfloat[]{\includegraphics[height=0.25\textwidth]{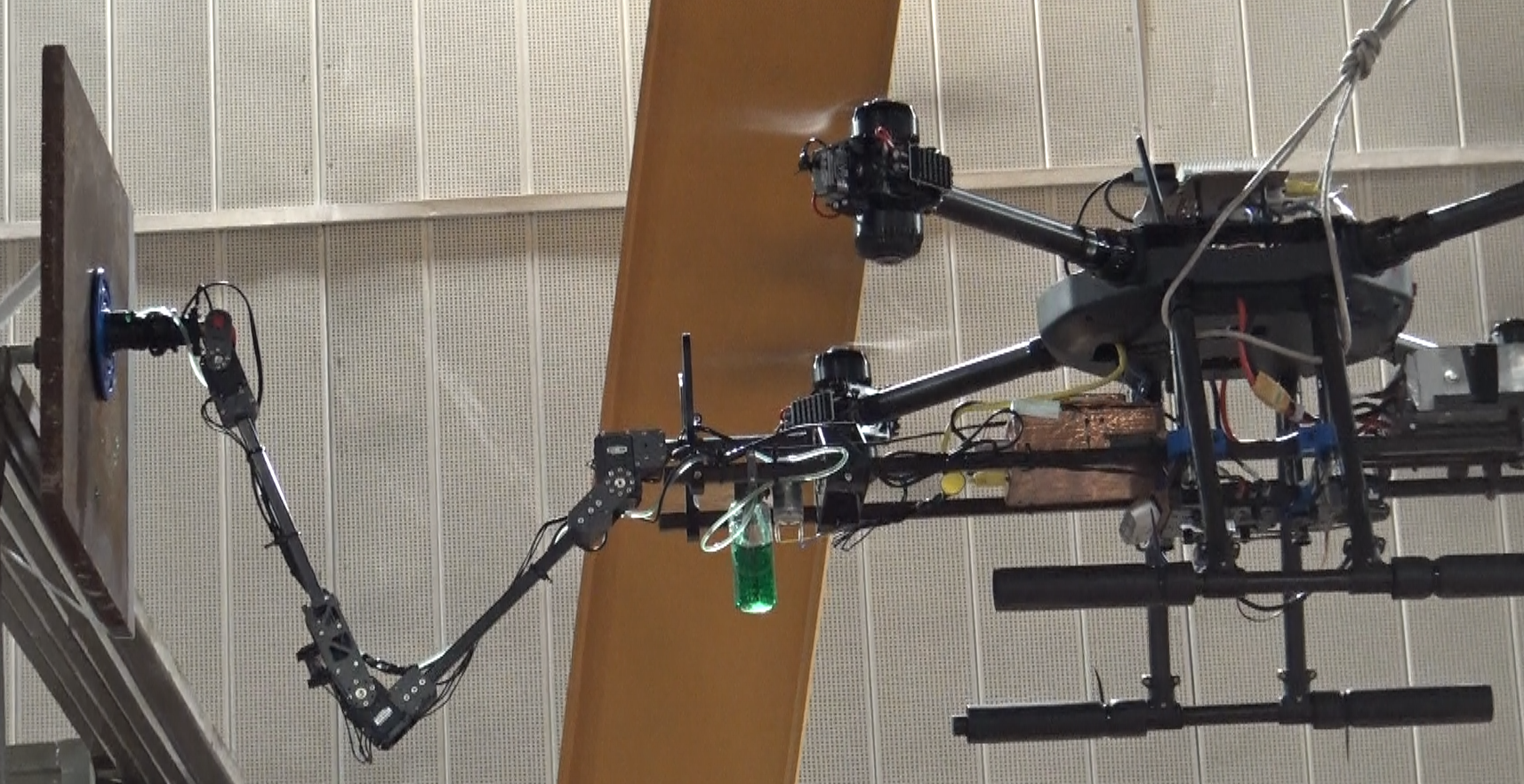}}
\caption{Proposed coaxial tilting octa-rotor equipped with a robotic arm. Snapshot in laboratory (a) and during task execution (b).}
\label{fig:drone}
\vspace{-0.5cm}
\end{figure}
In this context, the use of aerial platforms has been investigated as well. The actual commercial solution to remotely perform NDT measurements using aerial robots relies on telescopic tools attached to the robot frame. The APPELLIX~\cite{APPELLIX}, Texo Drone Survey Inspection platform~\cite{TEXO} and Ronik Inspectioneering UT device~\cite{RONIK} represent the most promising business technology in this field. \\
The main problem of these solutions is the stability of the floating base, which can be compromised when the robot is flying in contact with the inspection point. Therefore, to accomplish this type of task, the drone needs good tracking of the force that it exerts on the surface. For that purpose, like in \cite{10156642}, multi-rotors with tilted or tilting thrusters can be exploited thanks to their ability to exert a force in a desired direction, maintaining a desired orientation of the whole robot. The ability to preserve the stable hovering at different orientations is shown in \cite{voliropaper, Ryll, brunner_tilt} where the focus is given on both the drone design and the control allocation technique.\\
While tilting UAVs with fixed probes can achieve excellent results on flat surfaces like in \cite{brunner_hybrid} or obstacle-free spaces, they lack dexterity, in the case of pylons, T-shaped beams, and other typical structures of an industrial environment. On another side, one of the crucial points in the NDT for inspection and maintenance is to ensure the measurement's quality and consistency. For this purpose, only the long-term experience of specialized operators in the field can be effective. The combined use of a tilting aerial platform with a fully actuated robotic arm aims therefore to remotely allow the same dexterity to NDT operators as if they were on the spot.\\
This letter proposes a semi-autonomous system to perform the discussed inspection. The proposed platform, in fig.~\ref{fig:drone}, is composed of an omnidirectional drone and an actuated robotic arm carrying an inspection tool to complete the assigned task. The platform has to navigate the industrial facility and reach the inspection spot; then, a remote pilot triggers the different manipulator control modalities directly from the radio controller (RC). In the first step, the arm reaches a "home" position, ready to go in interaction; then, the switch between contact-free and contact-based control is handled by an impedance controller smoothly carrying the end-effector (E-E) in contact with the surface; finally, with a parallel force-impedance control it applies a desired force while maintaining the fixed hovering position. When the task is completed the arm returns to the "home" position decreasing the desired interaction force until deactivating the force control. The remote pilot can now repeat the measures in a different spot or command the drone to land and complete the task.\\
In the following section, both the platform and the control strategy will be presented showing an experiment in a real-world indoor scenario collecting all the results and some insight for future developments.

\section{Technical approach}
\begin{figure}[t]
\centering
\includegraphics[width=0.7\textwidth]{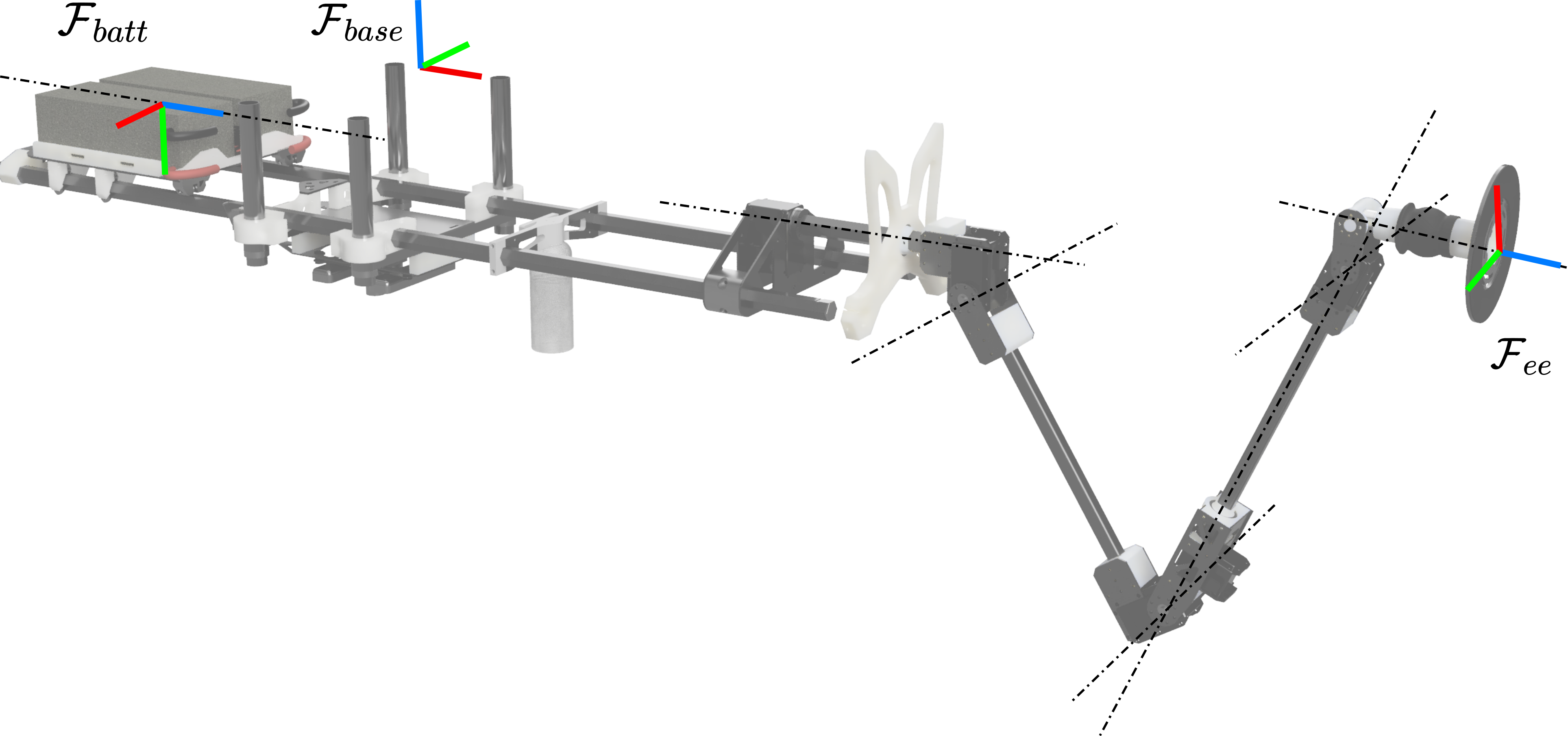}
\caption{Kinematic structure of the five DoF arm and moving battery carriage.}
\label{fig:arm_kinematics}
\vspace{-0.5cm}
\end{figure}
\subsection{Proposed platform}
The endowed drone features a coaxial tilting octa-rotor design, with each pair of co-axial motors linked to an independent servo-motor. It was developed by customizing a commercial frame and controlled through a standard Pixhawk flight controller. The PX4 firmware was modified without using third-party tools like the MATLAB toolbox as in~\cite{bird_div}. This improvement helps to preserve the original functionality of the PX4 flight control stack adding new capabilities. This letter starts from the results of~\cite{10156642} where a custom PX4 flight controller capable of controlling omnidirectional tilting drones is presented.\\
The drone is equipped with an ultra-lightweight robotic arm featuring five Degrees of Freedom (DoFs), accompanied by a 6-axis force and torque sensor mounted on the tip of the arm just before the end-effector. The kinematic structure of the arm is illustrated in fig.~\ref{fig:arm_kinematics}.
This configuration allows for achieving a high level of dexterity, enabling the system to accomplish NDT measurements in situations and places that are difficult to reach for other systems present on the market.
The proposed robotic arm weighs approximately 1.5 kg while having a payload capacity of 1 kg. This payload capacity, small compared to other robotic arm solutions present in the market, is enough for carrying the measuring probe and allows the arm to be extremely lightweight and not overload the UAV.\\
A commercial echometer acquires and processes data coming from a piezo-electric probe located inside the E-E, and a flange is placed on the tip to accommodate the surface that will be measured. Changing the shape of the flange, the arm can easily be adapted to different surfaces and curvatures.\\
With its motion, the arm moves the center of mass of the system: especially when fully extended, this can compromise the balancing of the UAV, increasing the stress on propellers and servomotors.
This behavior can be compensated by moving the batteries to maintain the balance of the system and keep the center of mass as close as possible to the center of the UAV. In particular, the batteries, with a total mass of 1.75 kg, are placed on a belt-driven carriage that moves the batteries of a displacement $x_{batt}$ given by the equation~\eqref{eq:disp_batt}
\begin{equation}
    x_{batt} = x_{0,batt} +  K \,x_{g},
    \label{eq:disp_batt}
\end{equation}
where $x_{batt}$ is the desired position of the batteries w.r.t. the geometric center of the drone, $x_{0,batt}$ is the offset from the initialization, $K$ is a properly tuned gain, and $x_{g}$ is the position of the arm center of mass along the $X$-axis evaluated with the dynamic model of the arm. All the arm and battery carriage control systems are implemented on an onboard microcontroller; also, a custom acquisition board is used to read and amplify data from the force and torque sensor. The system can communicate with the onboard PC through an Ethernet connection.

\subsection{Mathematical formulation}
\begin{figure} [t]
\centering
\includegraphics[trim={0.0cm 14.0cm 15.0cm  0.0cm},clip,width=0.7\columnwidth]{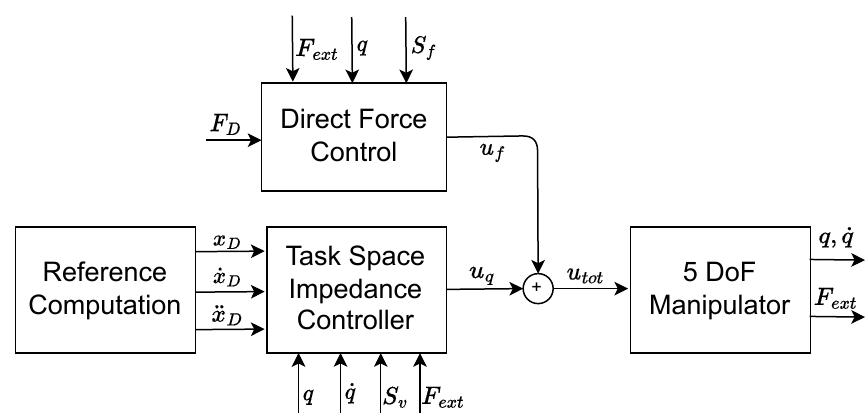}
\caption{Overall manipulator control architecture. The arm control inputs are computed by double integrating the auxiliary input. The matrices $S_f$ and $S_v$ select the Cartesian-space axis along which performs an impedance control or force control.}
\label{fig:inv_dyn}
\vspace{-0.5cm}
\end{figure}

Considering figure~\ref{fig:arm_kinematics}, let $\mathcal{W}$ be the fixed reference frame. Let $\mathcal{F}_{base}$ be the frame attached to the aerial manipulator's center of mass (CoM). Without loss of generality, at the beginning of the task, $\mathcal{W}$ and $\mathcal{F}_{base}$ are coincident. Let $\mathcal{F}_{ee}$ be the frame attached to the arm's E-E. All the mathematics formulations written below are defined in the $\mathcal{F}_{base}$. \\
The whole task should be executed in quasi-stationary flight conditions; in this situation, it is possible to consider the two robotic systems as decoupled. Also, the proposed control architecture is decoupled to achieve robustness and scalability of the whole system and implemented on the onboard microcontroller.\\
The flight controller PX4 custom firmware stabilizes the platform and ensures position tracking for precise measurement patterns, while the arm controller provides safe interaction with the inspected environment with a parallel force-impedance control~\cite{803301}. 
In our customized version of PX4~\cite{10156642}, the controller is the one already present in the autopilot, consisting of a cascade of P, PD, and PID controllers. The allocation matrix, instead, has been changed with the one proposed in~\cite{VOLIRO}, allowing to compute also the angle setpoint for the servomotors.\\
The E-E pose is controlled in the task space with an inverse Jacobian approach, ensuring the rejection of all external disturbances. A parallel impedance-force control is developed to ensure motion and force tracking simultaneously. In fig.~\ref{fig:inv_dyn}, the manipulator control architecture is presented: the probe's desired position, given by an operator or a measurement pattern, is filtered by an impedance control that uses the 6D force sensor feedback to achieve the desired compliancy on each direction while the probe is reaching a stable contact with the measured surface, preventing bumps to be propagated to the aerial platform. 
An impedance control law is developed to handle the transition between contact-less ($F_{ext}=0_6$) and contact-based motion ($F_{ext}\neq0_6$) correcting the Cartesian space references on the basis of the force feedback. Once the contact is established, the system switches to a direct force control on the approach direction to give adequate pressure for the NDT equipment to operate, while remaining compliant with the impedance control on the other directions. This selection is performed by the matrices
\begin{equation}
    \begin{matrix}
    S_v=\begin{bmatrix}
\begin{bmatrix}
0 & 0\\ 
1 & 0\\ 
0 & 1
\end{bmatrix} & \boldsymbol{0}_{3\times3}\\ 
\boldsymbol{0}_{3\times2} & \boldsymbol{I}_{3\times3}
\end{bmatrix} \in \mathbb{R}^{6\times n_v},   & & S_f=\begin{bmatrix}
\begin{bmatrix}
1\\ 
0\\ 
0
\end{bmatrix}\\ 
\boldsymbol{0}_{3\times1}
\end{bmatrix} \in \mathbb{R}^{6\times n_f}
    \end{matrix}.
\end{equation}
$n_v$ and $n_f$ represent the dimension of the velocity-controlled and force-controlled sub-spaces in the same way as~\cite{Siciliano}. 
This matrix is used in a proportional-derivative (PD) control scheme to track the desired force $f_{D}(t) \in \mathbb{R}^3$ nullifying the error 
\begin{equation}
    e_F(t)=\begin{bmatrix}
    f_{ext}(t)^T-f_{D}(t)^T \\\boldsymbol{0}_3^T
    \end{bmatrix}\in\mathbb{R}^{6}
\end{equation}
\begin{equation}
\label{pi_force}
u_f=J^\dagger(q)(I - S_vS_v^{\dagger})CS_f\Big(K_{P_f}S_f^{\dagger}e_F(t)+K_{D_f} \dot{\lambda_d}(t)\Big),
\end{equation}
where $C\in\mathbb{R}^{6\times6}$ is the compliance matrix of the spring between the two elements in contact, $K_{P_f}, K_{D_f}\in \mathbb{R}^{n_f\times n_f}$ are positive definite gains matrices and $\dot{\lambda_d}=S_f^{\dagger}\dot{e_F}\in\mathbb{R}$ the time derivative of the force error weighted by the force-controlled sub-space matrix. The exponent $(\cdot)^\dagger$ indicates the pseudo-inverse operator. The control scheme is completed by designing a Cartesian space impedance controller: taking into account the system dynamics, the joint torques are computed choosing a properly virtual control input $\ddot{q}=u_q\in \mathbb{R}^n$. Like in~\cite{mario}, the position set-points are computed by double integrating the auxiliary acceleration inputs $u_q$. Being $\tilde{x},\dot{\tilde{x}}\in \mathbb{R}^{6}$ the position and velocity errors, computed starting from the references and the actual feedback, it is then possible to impose
\begin{equation}\label{eq:INVDYN_OP_SPACE_stabi}
u_q = J^{\dagger} \Big(S_v\Big( M_a^{-1}(M_aS_v^{\dagger}\ddot{x}_D + K_{D_a}S_v^{\dagger}\dot{e}_m + K_{P_a}S_v^{\dagger}e - S_v^{\dagger}F_{ext}) - \dot{J}(q,\dot{q})\dot{q} \Big),
\end{equation}
where $M_a, K_{D_a}, K_{P_a} \in \mathbb{R}^{n_v \times n_v}$  represent the impedance mass, damping, and stiffness matrices, respectively; $F_{ext} = [f_{ext} \ m_{ext}]^T \in \mathbb{R}^6 $ is the measured wrench; The matrices $S_f$ and $S_v$, again, divide the 6D space into commands addressing the motion control task and others addressing the force control task. \\
Measuring the interaction force indicated with $F_{ext}$, the virtual control inputs can track the E-E's references computing the corresponding corrections being compliant with the interaction surface.
In the end, the parallel impedance-force control input is $u_{tot} = u_q + u_f$. This approach also ensures good measurement quality while the aerial platform is not perfectly stable in case of wind gusts that can be present for instance at height on the tank's surface. 

\section{Experiments completed or scheduled}
Several experiments have been performed both in simulation and in a real scenario consisting of an indoor GPS-denied environment proving the performances of the proposed setup. A mock-up of an industrial facility is considered for the interaction phase. The aerial platform feedback is computed starting from an onboard camera system: a combination of T265 and D435i RealSense cameras are considered. In complex robotics autonomous applications, simultaneous localization and mapping  (SLAM) algorithms are often necessary to obtain accurate and safe results. The IMU is subject to accumulative errors over time, meaning that the robot's position estimate becomes less accurate as time passes. SLAM, on the other hand, uses additional sensors to improve localization accuracy. In the actual case study, through RTABMAP, it was possible to map the indoor environment and retrieve the correct feedback info for the drone.\\
A human operator controls the drone movement on the map in the position flight mode sending the set-points through an RC toward the mock-up. While navigating, he can also send commands to the manipulator to change its behavior. The manipulator control sequence is composed of three phases: $(i)$ contact-free control keeping fixed the E-E pose in a "home" configuration; $(ii)$ approaching phase, triggering the impedance controller and starting the interaction being compliant with the environment; $(iii)$ contact-based control triggering the parallel controller applying a desired force on the surface while the echometer retrieve the correct measures. The different controller steps are triggered in sequence until the measure phase is completed returning to the already mentioned "home" configuration.\\
To obtain the thickness measures, it is needed to create a vacuum between the mock-up and the E-E: to this end, an electric pump is activated covering the E-E's tip with a viscous gel. To avoid undesired force measures after this phase, the sensor bias is nullified allowing to read the correct values from the sensors. 

\section{Main experimental insights}
\begin{figure}[t]
\centering
\includegraphics[trim={0.0cm 0.0cm 0.0cm  0.0cm},clip,width=0.8\columnwidth]{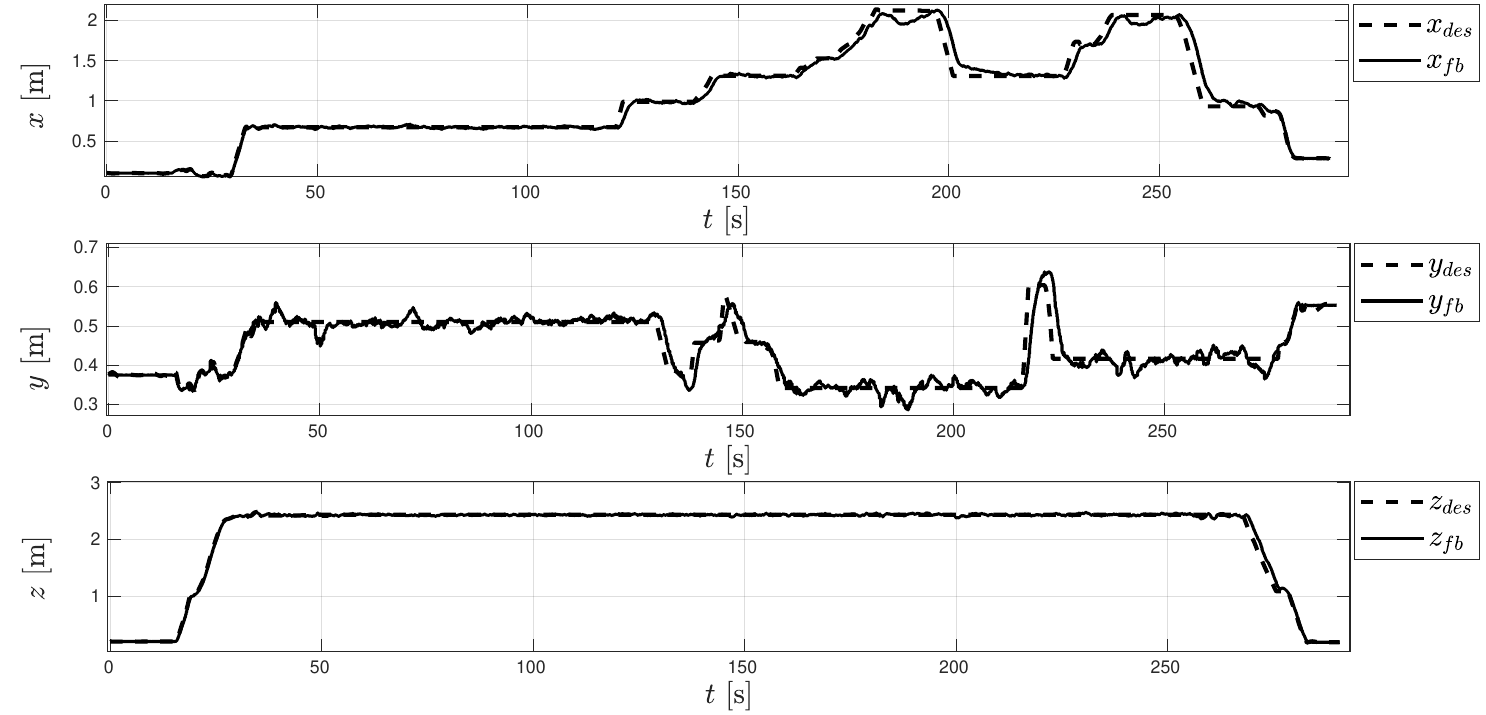}
\caption{Experimental results: drone's reference and actual linear position along the three Cartesian axes. The UAM counter-reacts to the external disturbance during the interaction preserving its pose by tilting the propellers.}
\label{fig:enu}
\vspace{-0.5cm}
\end{figure}

\begin{figure}[h]
\centering
\vspace{-0.5cm}
\includegraphics[width=0.8\columnwidth]{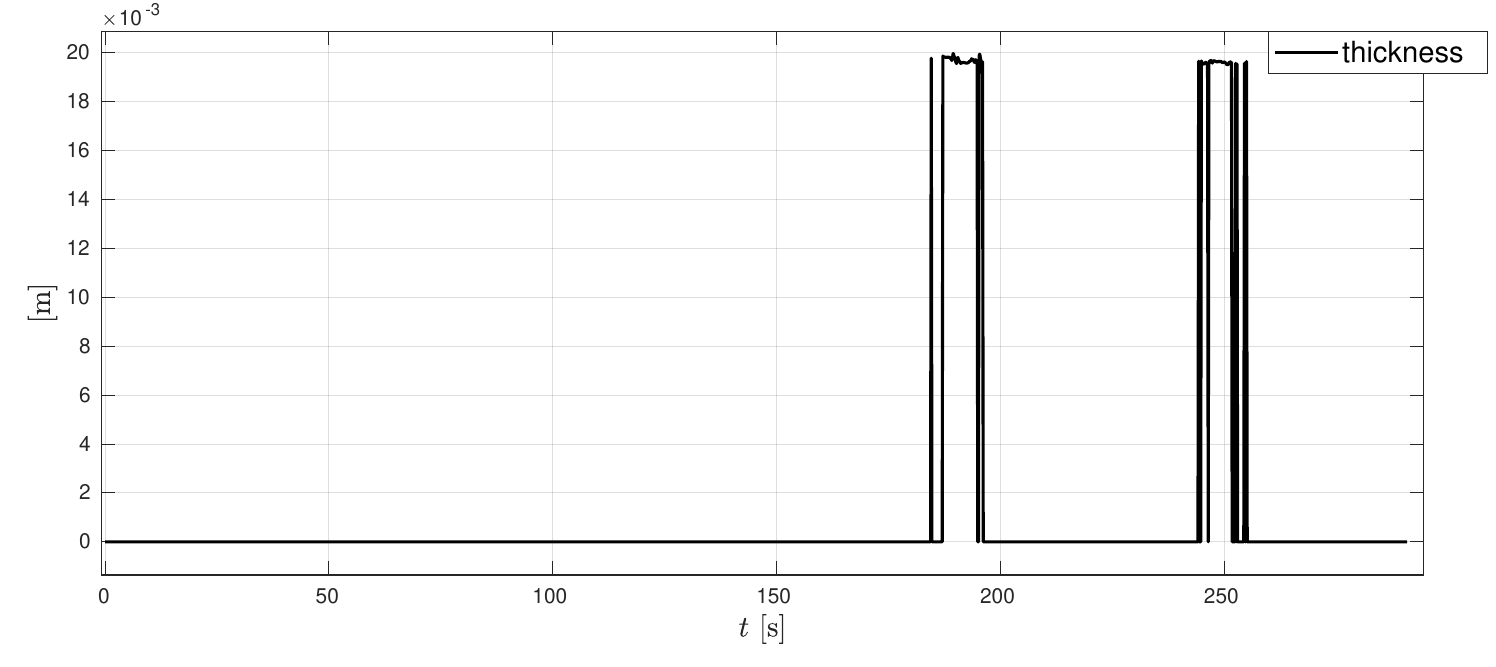}
\caption{Experimental results: echometer sensor feedback measuring the surface thickness during the interaction task. }
\label{fig:echo}
\vspace{-0.3cm}
\end{figure}

\begin{figure}[h]
\centering
\includegraphics[trim={0.0cm 0.0cm 0.0cm  0.0cm},clip,width=0.8\columnwidth]{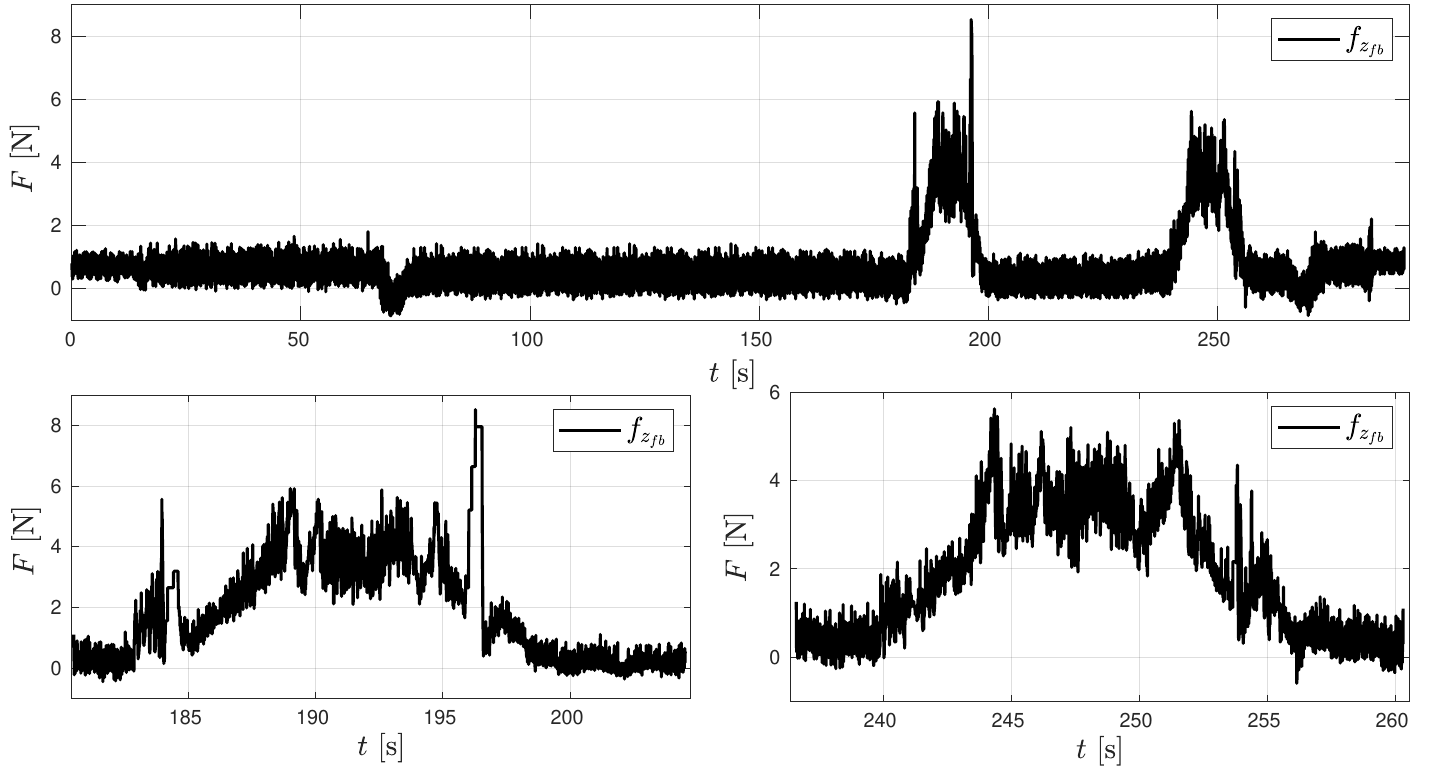}
\caption{Experimental results: (on the top) interaction force feedback with a zoom (on the bottom) in the two inspection phases. The force reference value is $3.5N$ in each try.}
\label{fig:force_contr}
\vspace{-0.3cm}
\end{figure}

In this section, the experimental results are presented showing the capability of the proposed platform. The completed test comprises two inspections and measures of the same interaction spot showing the repeatability and robustness of the proposed controller. The first phase of free flight in the GPS-denied environment shows the capability of the custom PX4 firmware to perfectly track the references. During the interaction, the drone stays still despite the external disturbances given by the manipulator completing the task.
%
Concerning the described controller, the impedance gains used for the experiments are: 
\begin{gather}
M_a = S_v^{-1}\begin{bmatrix}
0.5 I_{3\times3} & 0_{3\times3} \\
0_{3\times3} & 0.001 I_{3\times3}
\end{bmatrix}, \,
K_{P_a} = S_v^{-1}\begin{bmatrix}
30 I_{3\times3} & 0_{3\times3} \\
0_{3\times3} & 0.1 I_{3\times3}
\end{bmatrix}, \\
K_{D_a} = S_v^{-1}\begin{bmatrix}
7.0 I_{3\times3} & 0_{3\times3} \notag \\
0_{3\times3} & 0.2 I_{3\times3}
\end{bmatrix};
\label{eq:gains}
\end{gather} 
while the direct force control gains are:
\begin{gather}
K_{P_f} = S_f^{-1}\begin{bmatrix}
0.25 I_{3\times3} & 0_{3\times3} \\
0_{3\times3} & 0.5 I_{3\times3}
\end{bmatrix}, \,
K_{D_f} = S_f^{-1}\begin{bmatrix}
0.1 I_{3\times3} & 0_{3\times3} \\
0_{3\times3} & 0.1 I_{3\times3}
\end{bmatrix}.
\label{eq:gains2}
\end{gather} 
Figure~\ref{fig:enu} shows how the drone can easily track the desired references from the RC. The velocity references are integrated and used to compute the actual position references. The presence of the actuated manipulator creates some oscillation on the X-Y plane, but the flight controller can preserve the stable hovering. The tilting propellers allow the drone to remain still during the interaction.
\begin{figure} [h]
\centering
\includegraphics[trim={0.0cm 0.0cm 0.0cm  0.0cm},clip,width=0.8\columnwidth]{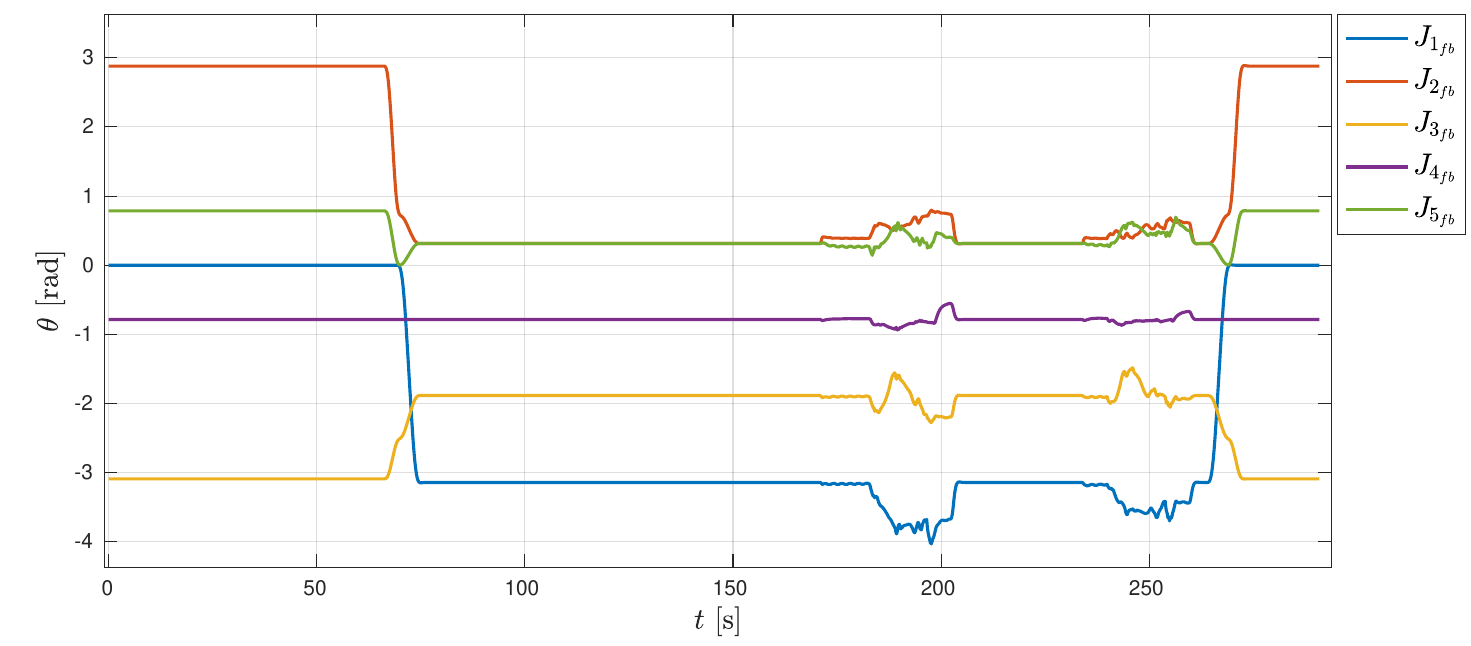}
\caption{Experimental results: arm's joints state feedback evolution. The impedance controller adjusts the joint angular position to be compliant w.r.t. the interaction surface.}
\label{fig:js}
\vspace{-0.5cm}
\end{figure}
%
The arm is actuated by five position-controlled servo motors, one for each joint. Its configuration changes during the task execution. From a "home" configuration the pilot can trigger the inspection phases when the drone reaches a stable hovering in the inspection spot nearby. The manipulator starts to approach the surface and the impedance control law adjusts its E-E position to establish a stable contact and continues to preserve it during the next pushing phase as can be appreciated in fig.~\ref{fig:js}.\\
In this experiment, two successive measure tests are performed on the same spot to show the controller behavior. As can be appreciated in fig.~\ref{fig:echo} and in fig.~\ref{fig:force_contr} the arm is capable of completing these tasks with high accuracy. The desired force is chosen equal to $F_D = 3.5N$: to create the vacuum. a small amount of force has to be exchanged by the systems. In fig.~\ref{fig:force_contr} $f_{z_{fb}}$ is the interaction force feedback component in the approaching direction converging to $F_D$ during the measures. During the measures, it is possible to see the accuracy of the echometer sensor showing the same results each time the surface is approached: in this case, the measured thickness is $\simeq 0.019m$.
\section{Results}
The paper proposes a novel solution to complete inspection tasks with a tilting aerial platform in an industrial environment. The results show the task's feasibility and the performance of the developed controller. It is possible to perform different measures during the flight without losing the controller's performance. The whole task is planned in an indoor environment. Some problems arise during outdoor flights: the principal causes can be found in the mapping and localization algorithms. Future works include investigating other ways to retrieve the correct drone feedback in outdoor scenarios considering the use of sensors like lidar or GPS. 
The results are retrieved in semi-autonomous flights. The principal updates to this work will be to remove the human operator from the control loop and make the whole system fully autonomous, with a coupling in the task and trajectory planning. Image elaboration techniques might also be introduced to detect notable spots on the interaction surfaces in parallel vision/force control algorithms.

%
%

\bibliographystyle{abbrv} 
\bibliography{paperbib.bib} 
\end{document}